\documentclass[runningheads]{llncs}

\usepackage[T1]{fontenc}
\usepackage{times}
\usepackage{url}
\usepackage{latexsym}

\usepackage{amsmath,amssymb,amsfonts}
\usepackage{algorithmic}
\usepackage{listings}
\usepackage{subfig}
\usepackage{multirow}
\usepackage{graphicx}
\usepackage{textcomp}
\usepackage{booktabs}
\usepackage{hyperref}
\usepackage{lstautogobble}
\usepackage{xcolor}
\usepackage{float}

\usepackage{color}

\begin{document}

\title{JaccDiv: A Metric and Benchmark for Quantifying Diversity of Generated Marketing Text in the Music Industry}
\titlerunning{JaccDiv}
\author{Anum Afzal\inst{*}\orcidID{0000-0001-8146-1949} \and
Alexandre Mercier\inst{}\orcidID{0009-0009-6483-2758} \and
Florian Matthes\inst{}\orcidID{0000-0002-6667-5452}}
\authorrunning{Afzal et al.}
%
\institute{Technical University of Munich,  Boltzmannstraße 3, 85748 Garching bei München, Germany\\
\email{\{anum.afzal\}@tum.de}} 

\maketitle
\begin{abstract}
Online platforms are increasingly interested in using Data-to-Text technologies to generate content and help their users. Unfortunately, traditional generative methods often fall into repetitive patterns, resulting in monotonous galleries of texts after only a few iterations. In this paper, we investigate LLM-based data-to-text approaches to automatically generate marketing texts that are of sufficient quality and diverse enough for broad adoption. We leverage Language Models such as T5, GPT-3.5, GPT-4, and LLaMa2 in conjunction with fine-tuning, few-shot, and zero-shot approaches to set a baseline for diverse marketing texts. We also introduce a metric JaccDiv to evaluate the diversity of a set of texts. This research extends its relevance beyond the music industry, proving beneficial in various fields where repetitive automated content generation is prevalent.
\end{abstract}


\section{Introduction}
\label{sec:introduction}
In an age where digital transformation is reshaping industries and artificial intelligence is seemingly introduced into every aspect of life, online platforms are adapting their services to ensure a smoother user experience. As most platforms' attractiveness relies on user-generated content, there is a need to provide users with the tools to create relevant content. LLM-driven technologies can be leveraged to generate text of sufficient quality for mass adoption \cite{brown_language_2020}, \cite{wei2022emergent}. However, controlling the quality, richness, and most importantly diversity of generated content is important. Diversity is crucial in the context of marketing on platforms, where competing products and services are often listed side-by-side and where descriptions for said products or services are usually the most distinguishing factor between competitors. In this case, the repetition of similar texts becomes apparent and can lead to a loss of trust in the platform. ~\cite{jentzsch_chatgpt_2023} show that for the complex domain of humor, ChatGPT repeats variations of the same 25 jokes, which is a high enough number to fool users into thinking that each joke is unique, but probably not diverse enough when marketing descriptions are generated for a platform with possibly hundreds or thousands of competitors. 

We worked with an online platform aimed at connecting musicians with event organizers. The platform allows the band to create their profile which typically includes a self-description. Since writing about themselves is a time-consuming process, many bands postpone this step or write a short text. To make the band's profile more appealing, an LLM-driven virtual assistant could help, by generating a description based on the data provided by the band during sign-up. Most studies so far have concentrated on reducing repetitions or controlling text diversity inside a single generated sample. In contrast, we aim to avoid similar structures and increase text diversity between samples. By comparing and contrasting various approaches, we identified the most effective techniques for achieving diverse and engaging marketing content within the music industry context. Our main contributions are as follows:

\paragraph{\textbf{Limited Dataset:}} As in any industrial use case, the dataset is noisy, limited, and sparse, so many samples do not contain complete band information. We show how to generate good-quality descriptions while leveraging such a dataset.
\paragraph{\textbf{Diversifying Content:}} Since the input might be similar for many bands, their output is expected to contain repetitions. We experiment with various techniques to increase the diversity of generated text.
\paragraph{\textbf{Text Evaluation:}} While there are many metrics to evaluate the quality of the generated text, metrics to evaluate diversity are limited. We proposed a new reference-free metric that can be used to measure the diversity between a corpus of texts.


\section{Related Work}

\label{chapter:relatedwork}
Over the years, many different approaches have been proposed for data-to-text generation. ~\cite{kasner-dusek-2022-neural} proposed a zero-shot approach by using multiple pre-trained models such that first the input triplet is formed into facts that are then ordered using a variation of BART-base. RoBERTa is then applied to decide which facts can be aggregated with the last step using BART-base again to aggregate and compress the paragraphs. State-of-the-art approaches however tend to use end-to-end solutions such as GPT models or T5 either in zero-shot mode~\cite{axelsson2023using} or in combination with additional steps such as disambiguation~\cite{xiang2022asdot}, reasoning~\cite{saha2022murmur} or Chain-of-Thought prompting~\cite{wei_chain--thought_2023,kojima_large_2023} or even Tree of Thought reasoning~\cite{yao_tree_2023}. Although these methods do try to achieve diverse texts, their research focus is concentrated on the quality of the text and controllability \cite{zhang_survey_2022} of the models. Some approaches such as ~\cite{elder-etal-2018-e2e}try to tackle this problem by adding relevant words that a model is then supposed to include. Some prompts modification approaches are discussed by ~\cite{puranik-etal-2023-protege,zhou2021informed}. For the specific case of question generation, ~\cite{cho2019mixture} achieved diverse questions by using multiple selectors that apply a mask on the input data, such that the model can focus on different parts of the input. In the context of product descriptions, tools like Alibaba's Luban \footnote{\href{https://www.alibabacloud.com/blog/alibaba-luban-ai-based-graphic-design-tool_594294?spm=a2c65.11461478.0.0.203f5355Rip1bW}{https://www.alibabacloud.com/luban}} use AI to generate graphical banners, and a common practice for large international commercial platforms is to use automatic translations in descriptions. Since our research will generate descriptions for users of a platform, the closest topic in marketing is generating descriptions for products in e-commerce~\cite{10.1145/3308558.3313407,10.1145/3292500.3330725}. Different approaches for diversity evaluation have been tried over the years \cite{liu_assessing_2013,tevet_evaluating_2021} utilizing various embedding techniques to evaluate content diversity. Lastly ~\cite{montahaei_jointly_2019} attempts to combine quality and diversity measurements into a single score.

\section{Dataset}
\label{sec:dataset}

We worked with an Industrial dataset acquired from the platform aimed at connecting musicians and event organizers. The dataset contained personal and public information about individual musicians grouped in formations (bands), event organizers who will interact with and hire the formations, and information about the events. We processed this raw dataset to construct a Formation dataset contains the Name, Type, Description, Homebase, Radius (distance they are willing to travel to), Genre and Event types.

\subsection{Data Preprocessing \& Filtering}
\label{subsec:data_preprocessing}

\begin{figure}[]
  \centering
  \includegraphics[width=\textwidth]{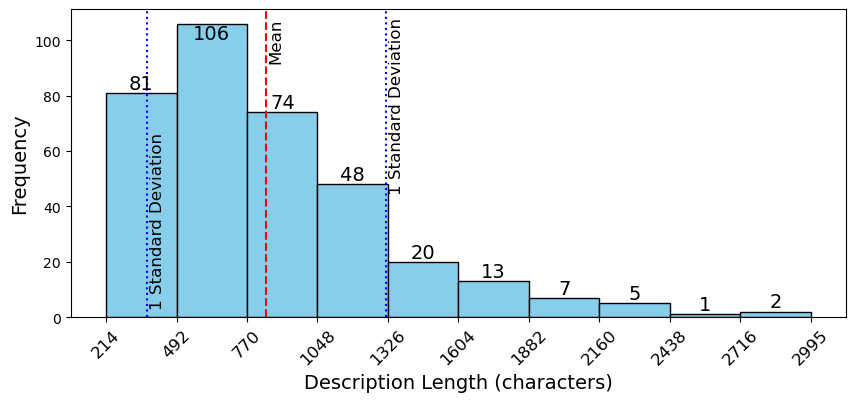}
  \caption{Distribution of description lengths.}
  \label{fig:lengthhistogram}
\end{figure}

 Since the dataset\footnote{The dataset was filtered to remove all the formations without description leaving 440 out of the 880 samples.} was in German, the descriptions were translated into English using the \href{https://www.deepl.com/en/pro-api}{DeepL API}. A distribution over the number of words in each description of the formation is shown in \autoref{fig:lengthhistogram}. The shortest text is only 214 characters long, and the longest is over 3000 characters long. The average length is about 863 characters with a standard variation of 512 characters. Due to some models' limited attention windows and for comparability, we will try to choose texts of similar lengths in the evaluation set described below.

\subsection{Manual Curation \& Quality Control}
All the remaining descriptions\footnote{The dataset was further manually filtered to remove unusable \& duplicate descriptions leaving 359 samples.} were manually rated in terms of quality and uniqueness. 
We use these good descriptions as references or templates when generating new descriptions. This subset was used for fine-tuning models and also for few-shot generation.

  

\section{Benchmark}

\subsection{Models}
\label{subsec:models}

We evaluate five (Large) Language Models in our experiments in terms of Quality and Diversity. We fine-tune T5-small ~\cite{raffel_exploring_2023}  \& Flan-T5-Base (250M parameters) ~\cite{chung2022scaling} through pre-fix tuning using input triplets. We use fine-tuned models as our baseline. We further experiment with GPT-3.5, GPT-4, LLama2-13b ~\cite{touvron_llama_2023} through the various diversity-enhancing techniques discussed below. We use GPT-3.5\footnote{\href{}{GPT-3.5-turbo-1106}} and GPT-4\footnote{\href{}{GPT-4-turbo-1106}} via the provided API and run LLama2-13b through llama.cpp\footnote{\href{https://github.com/ggerganov/llama.cpp}{https://github.com/ggerganov/llama.cpp}}, a framework for running large models on CPUs efficiently.

\subsection{Evaluation}\label{fund:evaluation}

To compare and ascertain the diversity between models and experiments, we selected a dataset of 50 bands. They will be used as a benchmark for different experiments and models in order to evaluate their quality and, most importantly, diversity. 

\paragraph{\textbf{Quality of Generated Text:}}
The reference descriptions in our dataset were written by humans, who had access to more information about their own band than we did. Usually, they did not include all the database information in their texts because parts of that data were displayed elsewhere on their profile. This makes reference-based evaluation such as ROUGE quite challenging and hence, we did not include any reference-based metrics in our evaluation. Instead, we leveraged LLM-based metrics~\cite{ke_ctrleval_2022,kocmi2023large} to achieve scores with better human correlation\cite{afzal-etal-2024-towards}. We choose \textbf{G-Eval}~\cite{liu_g-eval_2023}, a chain-of-thought driven \cite{wei_chain--thought_2023} prompt-based metric that uses GPT-4. The aim of a marketing text is to capture the audience's interest and attention in the hopes of ing a desired behavior. Therefore, we evaluate the texts as per the following features

\paragraph{\textbf{Engagingness:}} measures the ability of a text to engage a reader and judges how interesting a text might be from the way it is written\cite{mehri_usr_2020}.
\paragraph{\textbf{Fluency:}} measures the grammatical correctness of a text.\cite{fabbri_summeval_2021}
\paragraph{\textbf{Naturalness:}} used to determine whether the utterance could plausibly have been produced by a human.\cite{zhong_towards_2022}. 
 \paragraph{\textbf{Informativeness:}} used to determine whether the utterance contains all the information in the given content\cite{zhong_towards_2022}.

\paragraph{\textbf{Diversity in Text Generation:}}
\label{fund:diversity}
In contrast to quality evaluation where each sample is evaluated individually, diversity entails that multiple samples are compared to each other. The aim of our diversity analysis is essentially to quantify repetitions in a set of generated texts. It allows us to compare the diversity achieved within a group of texts generated by the model. Since there aren't many metrics that evaluate the diversity, we adapted the existing \textbf{Jaccard-similarity} coefficient, an n-gram-based method for comparing two texts. The Jaccard coefficient measures similarity between sample sets, here n-grams. It is calculated by dividing the size of the intersection of the two sets by the size of the union of the two sets. The coefficient is symmetrical, meaning the order of the texts doesn't matter, it is quick to compute and the result is a score between 0 and 1, with lower values signifying more diverse texts. By obtaining the pairwise Jaccard similarity of all generated samples in a set, we can get an average similarity score. We leverage the Jaccard Similarity to calculate the diversity as follows:
\[Diversity(U,V) = 1 - \frac{|U \cap V|}{|U \cup V|}\]

\begin{figure*}[h]
\small
  \centering
  \includegraphics[width=0.95\textwidth]{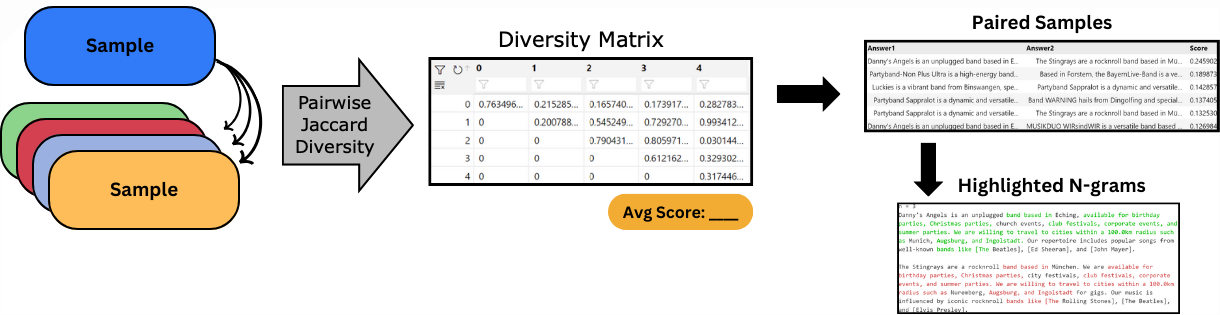}
  \caption[Evaluation Pipeline: Quality]{Evaluation Pipeline for our Diversity metrics}
  \label{fig:divPipe}
\end{figure*}

 A visual representation of the algorithm used for the diversity analysis is depicted in \autoref{fig:divPipe}. We generate bi-, tri-grams, ... n-grams for each text from a parameter n. The next step then compares all texts to each other by calculating the pairwise Jaccard similarity from the n-grams. The resulting values are stored in an upper diagonal matrix, as the similarity is symmetrical. From there, we can average all the values to get an approximate diversity score. We note, that the Jaccard similarity metric does not properly account for text length, but during development, all texts inside an experiment were generally of similar length, such that the impact of text length on the score was negligible.

\begin{figure}
  \centering
  \resizebox{\columnwidth}{!}{%
  \includegraphics[width=0.9\textwidth]{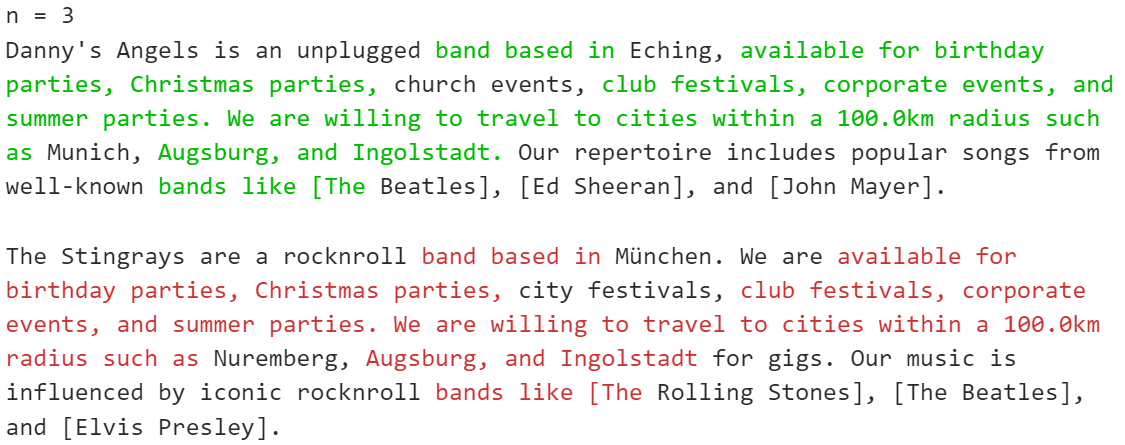}
  }
  \caption{For a given n, the n-grams found in both texts are highlighted. This method allows us to quickly identify reoccurring expressions or patterns.}
  \label{fig:div_analysis}
\end{figure}

\paragraph{\textbf{Human Evaluation:}}
Since manual annotation is a tedious task, we utilized the semi-automatic approach illustrated in \autoref{fig:divPipe} to aid the human annotators. We selected 5 models using the automatic diversity metric as a heuristic. Two annotators evaluate batches created by 5 descriptions for each model, so 10 comparisons per batch. The descriptions stem from the same bands for each model. The bands were randomly chosen beforehand. Annotators were asked to annotate each two possible pairs within a batch as shown in \autoref{fig:div_analysis} and assign a diversity score. This highlighting of similar 3-gram approach made it easier for annotators to compare the texts in terms of syntactic diversity.

\subsection{Experiments}\label{imp:experiments}
We experimented with the following hyperparameters to maximize diversity while still maintaining the overall quality fo the generated descriptions.
\paragraph{\textbf{Triplet Input:}} Following the approach by~\cite{gardent-etal-2017-webnlg}, we use the data-to-text generation technique by asking (L)LM to generate a coherent text using data points as input.  
\paragraph{\textbf{Temperature:}}
\label{imp:sub:temperature}
We experimented\footnote{The experiment was not run for LLaMa2 because of resource limitations.} with temperature values between 0.0 and 2.0 for GPT3.5 and GPT4. As the temperature parameter adds some randomness to the generation process, we expect it to significantly impact diversity. 

\paragraph{\textbf{Top-p:}}
This parameter focuses the next prediction on the smallest subset of the vocabulary, where the cumulative probability is above 'p'. The advantage of Top-p is that it strikes a balance between randomness and coherence. 

\paragraph{\textbf{Logit Bias:}}\label{imp:logb}
The logit bias is not a single parameter but rather a list of tokens (not words!) for which the output probabilities can be positively or negatively influenced. The logit bias is a powerful tool to influence the output of an LLM, as it can be used to force the model to use or stop using certain words or phrases.  The bias has a range of -100 to 100. After each generation, the occurrences of each token are counted, and the biases for the top 100 most used tokens are modified. 

\paragraph{\textbf{Data Ordering:}}\label{imp:sub:data_ordering}
We observed that the order of the data in the input had a significant impact on the text generation. As an example, if the music genres were specified as \textit{Pop, Jazz, Lounge} the models would by default use the same order when describing each genre. This led to repetitive texts especially when the generated text was short. To counteract this, we shuffled the data but kept its structure before feeding it to the model.

\paragraph{\textbf{Alternative Instructions:}}\label{imp:sub:alt_instructions}
Changing the instructions in the model prompt should have an impact on the output. We therefore created a set of alternate instructions with the intention of replacing both the beginning and the end of our prompt. 

\paragraph{\textbf{In-context Learning:}}
\label{imp:sub:fewshot}
We used existing descriptions from the curated dataset as few-shot examples. This method relied on references that were translated to English through Deepl, which would lose some of the language-specific intricacies and expressions. The goal was to see whether the largely free LLMs would incorporate existing expressions into their output.

\section{Results and Discussions:}
We provide an overview of our experiments in \autoref{tab:all_scores} and discuss our findings below.
\label{sec:results}

\begin{table}[h]
\caption{Overall quality scores averaged over Engagingness, Fluency, Naturalness, Informativeness (Qual) by G-eval and scaled to  [0- 1] \& Jaccard Similarity based Diversity scores (JaccDiv) calculated by averaging the pairwise diversity scores [0-1] of the generated text, by model and experiment.}
\vspace{5pt}
\label{tab:all_scores}
    \centering
    \begin{tabular}{lccc}
        \toprule
        \textbf{Model} & \textbf{Technique} & \textbf{Qual} & \textbf{JaccDiv}\\
        \midrule
        \multicolumn{4}{c}{\textbf{Baseline}} \\
        \midrule
        German       & Original &0.778 & 0.999\\
        English      & Original  & 0.787 & 0.999\\
        \midrule
        \multicolumn{4}{c}{\textbf{Fine-tuned Language Models}} \\
        \midrule
            {T5} & {Triplet Input }&{0.315}& {0.984} \\ 
            {Flan-T5} & {Triplet Input }&{0.465}& {0.994} \\ 
         \midrule
        \multicolumn{4}{c}{\textbf{Instruct-tuned Large Language Models}} \\
        \midrule
          \multirow{7}{*}{GPT-3.5} & {Triplet Input} &{0.829}& {0.985} \\ 
                                            & Base Prompt &{0.845}& {0.964}\\
                                             & Shuffled Order&{0.840}& {0.974}\\
                                          &Alternate Instructions&{0.849}& {0.989}\\
                                            & Fewshot&{0.657}& {0.993}\\
                                            & Fixed Logit Bias &{0.804}& {0.996}\\
                                            &Adaptive Logit Bias  &{0.767}& {0.996}\\
        \midrule
        \multirow{6}{*}{GPT-4} & Base Prompt&{0.881}& {0.995} \\ 
                                   & Shuffled Order &{0.878}& {0.994}\\
                                   & Alternate Instructions &\textbf{0.886}& {0.994}\\
                                   & Fewshot &{0.320}& {0.999}\\
                                   & Fixed Logit Bias &{0.801}& \textbf{0.998}\\
                                   & Adaptive Logit Bias &{0.843}& \textbf{0.998}\\                           
        \midrule
        {LLaMa-2} & Base Prompt &{0.868}& {0.983} \\ 
                 \multirow{2}{*} {(13b-chat)}    & Shuffled Order &{0.876}& {0.987}\\
                                   & Alternate Instructions &{0.874}& {0.985}\\
        \bottomrule
    \end{tabular}

\end{table}

\subsection{Quality Evaluation using GPT-4}
As shown in \autoref{tab:all_scores}, the fine-tuned models such as T5-small and FlanT5-base were not able to produce good-quality descriptions. We suspect that this is caused by the low quality of training data due to translation, longer descriptions, and availability of only a few good samples. The underwhelming performance of models with few-shot examples solidifies our theory that the problem indeed lies with the dataset as the few-shot learning almost always leads to better performance.

Having initially experimented with the same Input Triplet as the fine-tuned, GPT-3.5 gave good enough results. Given the same prompt, GPT3.5 and GPT-4 performed roughly the same but surprisingly, LLaMa2 exceeded expectations in terms of overall quality given its size. Our findings are mostly in line with the literature and common rankings of LLMs~\cite{huggingfaceLMSysChatbot}, although we found LLaMa2 to be more capable than GPT3.5 for our use-case.

Additionally, We also looked at the individual scores provided by G-Eval. It is rumored that LLMs are often biased towards the text generated by LLMs. We observed a similar pattern in the evaluation done using GPT-4 as shown in \autoref{tab:fluency_nat}. The human-written descriptions in both English and German are consistently rated worse than machine-generated texts. In contrast, the machine-generated texts are all highly rated on these four features, achieving high scores in almost all cases. GPT-3.5 with a zero-shot approach generated good descriptions as per G-Eval with minor differences across models in terms of grammatical quality, fluency, and engagingness.

\begin{table}[]
\caption{Scores for all models were computed using 50 samples except for LLaMa2 which was computed on 20 samples.}

\label{tab:fluency_nat}
\centering

\begin{tabular}{lcccc}

\toprule
\textbf{Model} & \textbf{Fluency [1 - 3]}& \textbf{Naturalness [1 - 3]} & \textbf{Informativeness [1 - 4]} & \textbf{Engaginess [1 - 5]}\\
\midrule
   Original DE & 2.48  & 2.75  & 2.63 &3.54 \\
    Original EN & 2.42  & 2.74  & 2.68& 3.78\\
    GPT-3.5 & 2.74  & 2.98 & 2.96 & 3.65 \\
    GPT-4 & 2.94  & 2.98  & 3.00 & 4.00  \\
    LLaMa2 & 2.99  & 2.97  & 2.93 & 4.00 \\
    \toprule
    
\end{tabular}

\end{table}


\subsection{Diversity Analysis through Jaccard similarity}

We introduced and used an n-gram-based metrics utilizing Jaccard similarity to evaluate the diversity of the generated texts. Summarized in \autoref{tab:all_scores}, the scores highlight variations between the models in terms of how diverse the generated texts are. we discuss the implications below:


\paragraph{\textbf{Base Prompt:}} Through prompt, we asked the model to augment the input data such as asking to include nearby cities or possible bands and songs a band might cover. Compared to the original approach consisting of data-triplets, this approach surprisingly reduced diversity. This could be because given the same prompt, the output followed its structure and was thus less diverse. We show the final prompt shown in \autoref{add:prompts}.

\paragraph{\textbf{Temperature:}} To increase the diversity, we experimented with parameters such as the temperature which can have a moderate impact on the diversity of samples by introducing randomness. For the GPT models, modifying the temperature did not impact the quality, up to a threshold of around 1.6, \textit{where catastrophic randomness took over}. Since reducing its value didn't improve quality but significantly reduced diversity, we kept the default value for all following experiments

\paragraph{\textbf{ Logit Bias:}} To enhance diversity, we transferred information about previous generations to the upcoming one. The results dramatically increased the diversity of the output. We experimented with Logit Bias, which is a more direct way of controlling the output. A logit defines the final layer of an LLM which shows all possible words and their probabilities. The word with the highest probability is typically predicted by the model. Through logit bias, we can influence this final layer model by telling it to reduce the probability of previously generated words or not consider them at all. First, we experimented with a Fixed Logit Bias where we assigned a value of -50 to the top 100 most used tokens from the previous iterations. The Logit bias experiments had a strong effect on the diversity. However, in some cases, words would simply miss in the middle of sentences. Thus strong logit biases affect the quality of generation.

\begin{figure}
  \centering
\resizebox{\columnwidth}{!}{%
 \includegraphics[width=1\textwidth]{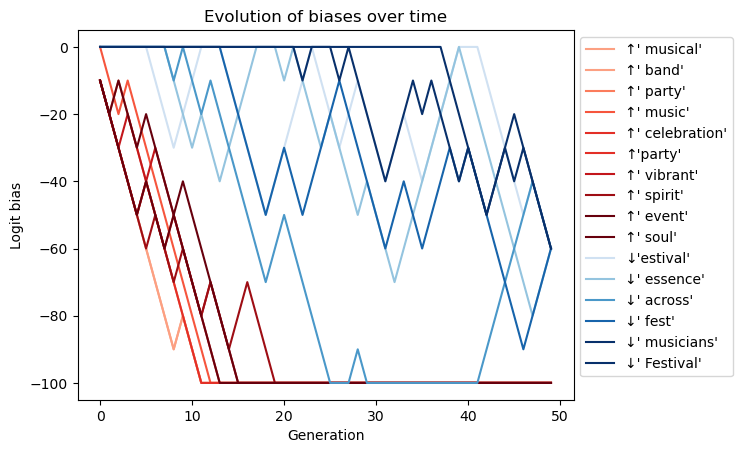}
 }
  \caption[Adaptive Logit Bias Values]{The evolution of the logit bias values for some tokens over 50 generations using an adaptive score. The most used tokens (red) did not get eliminated despite the biases quickly reaching their maximum value. Less used tokens among the top 100 (blue) were successfully limited, as shown by the fluctuating bias values.}
 \label{fig:logBScores}
\end{figure}

\paragraph{\textbf{Adaptive Logit Bias:}} 
Additionally, we used Adaptive logit bias where the bias value was derived based on how often a given token was used previously within the top 100 tokens. Shown in \autoref{fig:logBScores}, we observed that the most used tokens did not get eliminated despite the biases quickly reaching their maximum value. Less used tokens among the top 100 were successfully limited. We argue that Logit Bias is a good way to improve diversity but should be used sparingly and implemented carefully. 

\paragraph{\textbf{Shuffled Order:}} We experimented with shuffling the input data to observe if this would enhance the diversity of the generated text. This method had a significant impact on the diversity of GPT3.5 and LLaMa2. For GPT-4, the diversity seems to have been reduced slightly instead, although it is still significantly lower than for the other models. This could be because the models were already reordering parts of the input data while adding some text in between. These snippets did not change with shuffled input, they were merely reordered. All models had in common that shuffling the order of the input didn't impact quality. 

\paragraph{\textbf{Alternate Instructions:}}
Utilizing alternate instructions had a different impact on each model. Despite having the same meaning, the different formulations of the instructions were supposed to lead to other focus points and thus generate more diverse samples. In the case of LlaMa2, changing the first sentence in the prompt leads to differences in the output. Of course, each set of instructions had its own beginning, but at least it was possible to influence the model's output. For the GPT-3.5 model, the effect on diversity was very pronounced, having an increased effect compared to just shuffling. With GPT-4 it was once more the inverse, similar to the shuffling experiment.

\paragraph{\textbf{Few-shot Learning:}}Lastly, we experimented with few-shot techniques, by passing an existing band description with a simpler prompt. Unfortunately, few-shot learning leads to even worse results than the fine-tuned model. We suspect that this has something to do with the fact that the descriptions used as examples were also of lower quality as they were translated from German to English.

\subsection{Human Judgement}

Manual evaluation showed that LLM-generated texts were limited to data points, whereas many of the human-written texts included personal anecdotes or experiences that were not included in the structured data. Both fine-tuned models (T5 and Flan-T5) suffered from word looping. It should be noted that Flan-T5 was able to generate short band descriptions without any fine-tuning. Due to ambiguous naming in the input triplets and no context to rely on, they were unfortunately not usable. In most cases, the model would describe the band as being music genres/events types/etc., instead of using the proper verbs. GPT-4 generated longer but more hollow and less engaging texts. Whereas GPT-3 generated shorter texts and strictly adhered to the input data. A manual inspection showed that GPT-3.5 significantly benefited from shuffling the input order and changing the instructions slightly. Interestingly, GPT-4 was not influenced at all by these changes, and LlaMa2 was influenced much less than GPT-3.5. LlaMa2 regularly uses emojis, whereas GPT-4 only adds them when explicitly asked to do so. This shows that the models are not equally affected by the same changes. Given the same prompt, GPT-4 and LlaMa2 extrapolated more details, suggesting to a potential event planner why this band would be a good fit and what they could expect. We also noticed in manual evaluation and through our automatic scores that LlaMa2 as compared to other LLMs didn't perform so well in terms of diversity.
\paragraph{\textbf{JaccDiv correlation with Human Judgement:}}
For diversity evaluation, Two annotators manually annotated the same 50 pairs using the 2-gram highlighting techniques illustrated in \autoref{fig:div_analysis}). The Inter-annotator agreement between both annotators as per Cohen's kappa $\kappa$ [-1 to +1] is 0.214. Each model was evaluated on a subset of 5 bands. The average diversity scores calculated by JaccDiv and Human Annotators are shown in \autoref{tab:JaccDiv-coorelation}. It can be seen that Human Annotators are more strict when evaluating diversity as compared to an automatic score. Among all models, GPT4 with AdaptLogitBias is scored highest by both Human and JaccDiv score.

\begin{table}[]
\caption{Average Diversity scores calculated by JaccDiv and Human Annotators.}
\label{tab:JaccDiv-coorelation}
\centering

\begin{tabular}{lccc}

\toprule
\textbf{Model} & \textbf{Description} & \textbf{JaccDiv}& \textbf{Human}\\
\midrule
   Original & References in German& {0.993}& {0.990}   \\
    GPT-3.5 & Base Prompt& {0.926}& {0.320} \\
    GPT-4 &  Base Prompt&{0.982}& {0.705}\\
    GPT4& Alternate Instructions&  {0.986}& {0.690}  \\
    GPT4& Adaptive Logit Bias & {0.9903}& {0.885}  \\
    \toprule
    
\end{tabular}
\end{table}

\section{Conclusion}

Through our research, we have made significant contributions to the application aspects of LLMs in the Marketing Industry. We experimented with different models, methods, and input formats to see how each affected the quality and more importantly the diversity of generated descriptions. We found that randomness, induced by query shuffling, temperature parameters, and prompt modification, had different impacts on each model and was thus unreliable at improving the diversity. For a more consistent and reliable diversity-enhancing technique, we found that transferring previously used tokens to new prompts was a very effective, controllable, and reliable way. This technique has the drawback, that its scalability is limited by the amount of information a model can ingest. Future research on diversity should start with simpler datasets with have reliable references, and concentrate on the most effective way of transferring relevant information between prompts. We also found during our prompt engineering phase, that including CoT techniques in prompts was very effective at improving the quality and relevance of the output and augmenting the existent data with new insights, like geographical knowledge. On the flip side, this restricted the diversity of the output, as the model was boxed in with additional instructions. To tackle this tradeoff, we suggest researching ways to augment the data without restricting the model's creativity. Lastly, we introduced a simple but effective Jaccard-similarity metric for evaluating diversity which enabled us to compare the diversity of texts to each other, as well as entire datasets across models and techniques. We hope our insights will provide a good base for potential implementations of data-to-text systems in production environments, as well as inspire future research on the topic of diversity in Natural Language Generation.

\section*{Acknowledgements}
This project is supported by the German Federal Ministry of Education and Research (BMBF) grant 01IS17049 Software Campus 2.0 (TU München). 



\bibliographystyle{splncs04}
\bibliography{iciai}

\begin{thebibliography}{10}
\providecommand{\url}[1]{\texttt{#1}}
\providecommand{\urlprefix}{URL }
\providecommand{\doi}[1]{https://doi.org/#1}

\bibitem{huggingfaceLMSysChatbot}
{L}{M}{S}ys {C}hatbot {A}rena {L}eaderboard - a {H}ugging {F}ace {S}pace by lmsys --- huggingface.co. \url{https://huggingface.co/spaces/lmsys/chatbot-arena-leaderboard}, [Accessed 13-01-2024]

\bibitem{afzal-etal-2024-towards}
Afzal, A., Kowsik, A., Fani, R., Matthes, F.: Towards optimizing and evaluating a retrieval augmented {QA} chatbot using {LLM}s with human-in-the-loop. In: Dragut, E., Li, Y., Popa, L., Vucetic, S., Srivastava, S. (eds.) Proceedings of the Fifth Workshop on Data Science with Human-in-the-Loop (DaSH 2024). pp. 4--16. Association for Computational Linguistics, Mexico City, Mexico (Jun 2024). \doi{10.18653/v1/2024.dash-1.2}

\bibitem{axelsson2023using}
Axelsson, A., Skantze, G.: Using large language models for zero-shot natural language generation from knowledge graphs (2023)

\bibitem{brown_language_2020}
Brown, T.B., Mann, B., Ryder, N., Subbiah, M., Kaplan, J., Dhariwal, P., Neelakantan, A., Shyam, P., Sastry, G., Askell, A., Agarwal, S., Herbert-Voss, A., Krueger, G., Henighan, T., Child, R., Ramesh, A., Ziegler, D.M., Wu, J., Winter, C., Hesse, C., Chen, M., Sigler, E., Litwin, M., Gray, S., Chess, B., Clark, J., Berner, C., McCandlish, S., Radford, A., Sutskever, I., Amodei, D.: Language {Models} are {Few}-{Shot} {Learners} (May 2020)

\bibitem{10.1145/3292500.3330725}
Chen, Q., Lin, J., Zhang, Y., Yang, H., Zhou, J., Tang, J.: Towards knowledge-based personalized product description generation in e-commerce. In: Proceedings of the 25th ACM SIGKDD International Conference on Knowledge Discovery \& Data Mining. p. 3040–3050. KDD '19, Association for Computing Machinery, New York, NY, USA (2019). \doi{10.1145/3292500.3330725}, \url{https://doi.org/10.1145/3292500.3330725}

\bibitem{cho2019mixture}
Cho, J., Seo, M., Hajishirzi, H.: Mixture content selection for diverse sequence generation (2019)

\bibitem{chung2022scaling}
Chung, H.W., Hou, L., Longpre, S., Zoph, B., Tay, Y., Fedus, W., Li, Y., Wang, X., Dehghani, M., Brahma, S., Webson, A., Gu, S.S., Dai, Z., Suzgun, M., Chen, X., Chowdhery, A., Castro-Ros, A., Pellat, M., Robinson, K., Valter, D., Narang, S., Mishra, G., Yu, A., Zhao, V., Huang, Y., Dai, A., Yu, H., Petrov, S., Chi, E.H., Dean, J., Devlin, J., Roberts, A., Zhou, D., Le, Q.V., Wei, J.: Scaling instruction-finetuned language models (2022)

\bibitem{elder-etal-2018-e2e}
Elder, H., Gehrmann, S., O{'}Connor, A., Liu, Q.: {E}2{E} {NLG} challenge submission: Towards controllable generation of diverse natural language. In: Krahmer, E., Gatt, A., Goudbeek, M. (eds.) Proceedings of the 11th International Conference on Natural Language Generation. pp. 457--462. Association for Computational Linguistics, Tilburg University, The Netherlands (Nov 2018). \doi{10.18653/v1/W18-6556}, \url{https://aclanthology.org/W18-6556}

\bibitem{fabbri_summeval_2021}
Fabbri, A.R., Kryściński, W., McCann, B., Xiong, C., Socher, R., Radev, D.: {SummEval}: {Re}-evaluating {Summarization} {Evaluation} (Feb 2021). \doi{10.48550/arXiv.2007.12626}, \url{http://arxiv.org/abs/2007.12626}, arXiv:2007.12626 [cs]

\bibitem{gardent-etal-2017-webnlg}
Gardent, C., Shimorina, A., Narayan, S., Perez-Beltrachini, L.: The {W}eb{NLG} challenge: Generating text from {RDF} data. In: Alonso, J.M., Bugar{\'\i}n, A., Reiter, E. (eds.) Proceedings of the 10th International Conference on Natural Language Generation. pp. 124--133. Association for Computational Linguistics, Santiago de Compostela, Spain (Sep 2017). \doi{10.18653/v1/W17-3518}, \url{https://aclanthology.org/W17-3518}

\bibitem{jentzsch_chatgpt_2023}
Jentzsch, S., Kersting, K.: {ChatGPT} is fun, but it is not funny! {Humor} is still challenging {Large} {Language} {Models}. In: Proceedings of the 13th {Workshop} on {Computational} {Approaches} to {Subjectivity}, {Sentiment}, \& {Social} {Media} {Analysis}. pp. 325--340. Association for Computational Linguistics, Toronto, Canada (Jul 2023). \doi{10.18653/v1/2023.wassa-1.29}, \url{https://aclanthology.org/2023.wassa-1.29}

\bibitem{kasner-dusek-2022-neural}
Kasner, Z., Dusek, O.: Neural pipeline for zero-shot data-to-text generation. In: Muresan, S., Nakov, P., Villavicencio, A. (eds.) Proceedings of the 60th Annual Meeting of the Association for Computational Linguistics (Volume 1: Long Papers). pp. 3914--3932. Association for Computational Linguistics, Dublin, Ireland (May 2022). \doi{10.18653/v1/2022.acl-long.271}, \url{https://aclanthology.org/2022.acl-long.271}

\bibitem{ke_ctrleval_2022}
Ke, P., Zhou, H., Lin, Y., Li, P., Zhou, J., Zhu, X., Huang, M.: {CTRLEval}: {An} {Unsupervised} {Reference}-{Free} {Metric} for {Evaluating} {Controlled} {Text} {Generation}. In: Proceedings of the 60th {Annual} {Meeting} of the {Association} for {Computational} {Linguistics} ({Volume} 1: {Long} {Papers}). pp. 2306--2319. Association for Computational Linguistics, Dublin, Ireland (May 2022). \doi{10.18653/v1/2022.acl-long.164}, \url{https://aclanthology.org/2022.acl-long.164}

\bibitem{kocmi2023large}
Kocmi, T., Federmann, C.: Large language models are state-of-the-art evaluators of translation quality (2023)

\bibitem{kojima_large_2023}
Kojima, T., Gu, S.S., Reid, M., Matsuo, Y., Iwasawa, Y.: Large {Language} {Models} are {Zero}-{Shot} {Reasoners} (Jan 2023). \doi{10.48550/arXiv.2205.11916}, \url{http://arxiv.org/abs/2205.11916}, arXiv:2205.11916 [cs]

\bibitem{liu_assessing_2013}
Liu, H., Wang, P.: Assessing {Sentence} {Similarity} {Using} {WordNet} based {Word} {Similarity}. JSW  \textbf{8}(6),  1451--1458 (Jun 2013). \doi{10.4304/jsw.8.6.1451-1458}, \url{http://ojs.academypublisher.com/index.php/jsw/article/view/9168}

\bibitem{liu_g-eval_2023}
Liu, Y., Iter, D., Xu, Y., Wang, S., Xu, R., Zhu, C.: G-{Eval}: {NLG} {Evaluation} using {GPT}-4 with {Better} {Human} {Alignment} (May 2023). \doi{10.48550/arXiv.2303.16634}, \url{http://arxiv.org/abs/2303.16634}, arXiv:2303.16634 [cs]

\bibitem{mehri_usr_2020}
Mehri, S., Eskenazi, M.: {USR}: {An} {Unsupervised} and {Reference} {Free} {Evaluation} {Metric} for {Dialog} {Generation}. In: Proceedings of the 58th {Annual} {Meeting} of the {Association} for {Computational} {Linguistics}. pp. 681--707. Association for Computational Linguistics, Online (Jul 2020). \doi{10.18653/v1/2020.acl-main.64}, \url{https://aclanthology.org/2020.acl-main.64}

\bibitem{montahaei_jointly_2019}
Montahaei, E., Alihosseini, D., Baghshah, M.S.: Jointly {Measuring} {Diversity} and {Quality} in {Text} {Generation} {Models} (May 2019). \doi{10.48550/arXiv.1904.03971}, \url{http://arxiv.org/abs/1904.03971}, arXiv:1904.03971 [cs, stat]

\bibitem{puranik-etal-2023-protege}
Puranik, V., Majumder, A., Chaoji, V.: {PROTEGE}: Prompt-based diverse question generation from web articles. In: Bouamor, H., Pino, J., Bali, K. (eds.) Findings of the Association for Computational Linguistics: EMNLP 2023. pp. 5449--5463. Association for Computational Linguistics, Singapore (Dec 2023). \doi{10.18653/v1/2023.findings-emnlp.362}, \url{https://aclanthology.org/2023.findings-emnlp.362}

\bibitem{raffel_exploring_2023}
Raffel, C., Shazeer, N., Roberts, A., Lee, K., Narang, S., Matena, M., Zhou, Y., Li, W., Liu, P.J.: Exploring the {Limits} of {Transfer} {Learning} with a {Unified} {Text}-to-{Text} {Transformer} (Sep 2023). \doi{10.48550/arXiv.1910.10683}, \url{http://arxiv.org/abs/1910.10683}, arXiv:1910.10683 [cs, stat] version: 4

\bibitem{saha2022murmur}
Saha, S., Yu, X.V., Bansal, M., Pasunuru, R., Celikyilmaz, A.: Murmur: Modular multi-step reasoning for semi-structured data-to-text generation (2022)

\bibitem{tevet_evaluating_2021}
Tevet, G., Berant, J.: Evaluating the {Evaluation} of {Diversity} in {Natural} {Language} {Generation}. In: Merlo, P., Tiedemann, J., Tsarfaty, R. (eds.) Proceedings of the 16th {Conference} of the {European} {Chapter} of the {Association} for {Computational} {Linguistics}: {Main} {Volume}. pp. 326--346. Association for Computational Linguistics, Online (Apr 2021). \doi{10.18653/v1/2021.eacl-main.25}, \url{https://aclanthology.org/2021.eacl-main.25}

\bibitem{touvron_llama_2023}
Touvron, H., Martin, L., Stone, K., Albert, P., Almahairi, A., Babaei, Y., Bashlykov, N., Batra, S., Bhargava, P., Bhosale, S., Bikel, D., Blecher, L., Ferrer, C.C., Chen, M., Cucurull, G., Esiobu, D.: Llama 2: {Open} {Foundation} and {Fine}-{Tuned} {Chat} {Models} (Jul 2023). \doi{10.48550/arXiv.2307.09288}, \url{http://arxiv.org/abs/2307.09288}, arXiv:2307.09288 [cs]

\bibitem{wei2022emergent}
Wei, J., Tay, Y., Bommasani, R., Raffel, C., Zoph, B., Borgeaud, S., Yogatama, D., Bosma, M., Zhou, D., Metzler, D., Chi, E.H., Hashimoto, T., Vinyals, O., Liang, P., Dean, J., Fedus, W.: Emergent abilities of large language models (2022)

\bibitem{wei_chain--thought_2023}
Wei, J., Wang, X., Schuurmans, D., Bosma, M., Ichter, B., Xia, F., Chi, E., Le, Q., Zhou, D.: Chain-of-{Thought} {Prompting} {Elicits} {Reasoning} in {Large} {Language} {Models} (Jan 2023). \doi{10.48550/arXiv.2201.11903}, \url{http://arxiv.org/abs/2201.11903}, arXiv:2201.11903 [cs]

\bibitem{xiang2022asdot}
Xiang, J., Liu, Z., Zhou, Y., Xing, E.P., Hu, Z.: Asdot: Any-shot data-to-text generation with pretrained language models (2022)

\bibitem{yao_tree_2023}
Yao, S., Yu, D., Zhao, J., Shafran, I., Griffiths, T.L., Cao, Y., Narasimhan, K.: Tree of {Thoughts}: {Deliberate} {Problem} {Solving} with {Large} {Language} {Models} (May 2023). \doi{10.48550/arXiv.2305.10601}, \url{http://arxiv.org/abs/2305.10601}, arXiv:2305.10601 [cs]

\bibitem{zhang_survey_2022}
Zhang, H., Song, H., Li, S., Zhou, M., Song, D.: A survey of controllable text generation using transformer-based pre-trained language models. ACM Computing Surveys  (2022), publisher: ACM New York, NY

\bibitem{10.1145/3308558.3313407}
Zhang, T., Zhang, J., Huo, C., Ren, W.: Automatic generation of pattern-controlled product description in e-commerce. In: The World Wide Web Conference. p. 2355–2365. WWW '19, Association for Computing Machinery, New York, NY, USA (2019). \doi{10.1145/3308558.3313407}, \url{https://doi.org/10.1145/3308558.3313407}

\bibitem{zhong_towards_2022}
Zhong, M., Liu, Y., Yin, D., Mao, Y., Jiao, Y., Liu, P., Zhu, C., Ji, H., Han, J.: Towards a {Unified} {Multi}-{Dimensional} {Evaluator} for {Text} {Generation} (Oct 2022). \doi{10.48550/arXiv.2210.07197}, \url{http://arxiv.org/abs/2210.07197}, arXiv:2210.07197 [cs]

\bibitem{zhou2021informed}
Zhou, G., Lampouras, G.: Informed sampling for diversity in concept-to-text nlg (2021)

\end{thebibliography}
\end{document}